\documentclass[runningheads]{llncs}
\usepackage[hyphens]{url}
\usepackage[T1]{fontenc}
\usepackage{graphicx,verbatim}
\usepackage{booktabs}
\usepackage{subfig}
\usepackage{makecell}

\usepackage{color}
\usepackage[colorlinks=true, linkcolor=blue, citecolor=blue, urlcolor=blue]{hyperref}

\usepackage{colortbl}
\usepackage[usenames,dvipsnames]{xcolor}
\usepackage{pgfplots}
\usepackage{siunitx} 
\begin{document}

\title{A Systematic Benchmark of Intensity Normalisation Methods for 3D Knee MRI Segmentation and Cross-Domain Generalisability}

\titlerunning{MRI Normalisation and Knee Segmentation Generalisability}

\author{Oliver Mills\inst{1}\orcidID{0009-0001-9554-5788} \and
Philip Conaghan\inst{1,2}\orcidID{0000-0002-3478-5665} \and
Samuel Relton\inst{1}\orcidID{0000-0003-0634-4587}}
\authorrunning{Mills et al.}
%
\institute{University of Leeds, Leeds, UK \and
NIHR Leeds Biomedical Research Centre, Leeds, UK}

\maketitle              


\begin{abstract}
Robust out-of-the-box performance is essential for the clinical deployment of deep learning models in medical imaging. An important but underexplored factor affecting model generalisability is intensity normalisation, particularly for magnetic resonance imaging (MRI), where image intensities vary across scanners and protocols. In this study, we systematically compared seven normalisation methods and their impact on the performance of a 3D U-Net model for meniscus segmentation from knee MRI. The methods included standard scaling approaches, histogram-based techniques, and a Gaussian Mixture Model (GMM)-based method. Models were trained on the IWOAI 2019 dataset and evaluated on both internal and external test sets (SKM-TEA) to assess generalisability. Performance was similar internally but differences were significant on external data, with Z-score, Ny\'{u}l histogram matching, and CLAHE showing greater robustness than other methods. However, these differences were small compared to the significant performance drop observed between datasets. Overall, while intensity normalisation had a measurable effect on model generalisability, its impact was limited relative to the effects of domain shift, highlighting the need for complementary strategies for robust deployment.


\end{abstract}

\section{Introduction}
\label{sec:intro}
Segmentation is a key task in medical image analysis, enabling measurement of tissue boundaries and volumes. Deep learning models have shown strong performance in this domain, offering improved efficiency and consistency \cite{litjensSurveyDeepLearning2017}. However, they are sensitive to data characteristic changes and perform poorly on images that differ from their training data distribution \cite{kondratevaDomainShiftComputer2021}. This issue, known as domain shift, is common in medical imaging due to variations in scanner type, imaging protocol, and patient population. In MRI, signal variations occur even when the same scanner is used. Preprocessing and normalisation techniques help mitigate these effects by keeping image intensity distributions consistent, and in doing so stabilising training and improving model robustness \cite{ghazvanchahiEffectIntensityStandardization2024}. While fine-tuning models on target-domain data can help reduce the impact of domain shift, it is often infeasible due to the scarcity of accurate annotations in medical imaging. As a result, models are usually trained on data from select centres, highlighting the need for robust models that perform well on unseen institutions. Recent work has explored AI-based approaches for harmonisation and domain adaptation, including style transfer and adversarial methods, as well as statistical techniques such as ComBat \cite{johnsonAdjustingBatchEffects2007}. However, these approaches typically require access to target-domain data and additional training or fine-tuning, which may not be practical in many deployment settings. In contrast, the role of standard intensity normalisation methods not requiring further retraining in improving out-of-distribution robustness has received limited attention outside of brain MRI. This study addresses that gap by systematically comparing seven intensity normalisation methods for 3D meniscus segmentation in knee MRI.

Meniscus segmentation is important for quantitative assessment in osteo\-arthritis research in order to investigate the role the tissue plays in the disease pathway \cite{englundEffectMeniscalDamage2007,kornaatOsteoarthritisKneeAssociation2006,bhattacharyyaClinicalImportanceMeniscal2003,lenchikAutomatedSegmentationTissues2019,englundMeniscusPathologyOsteoarthritis2012}. However, it remains a challenging task due to the tissue's small size, irregular shape, and low contrast with surrounding tissues \cite{rahmanAutomaticSegmentationKnee2020}. These factors make accurate segmentation difficult, even for human experts, and increase model sensitivity to variations in image intensity and quality. Therefore, meniscus segmentation provides a useful test case for evaluating how intensity normalisation methods affect model robustness across diverse MRI data. While intensity normalisation has been studied in brain and breast MRI, its role in musculoskeletal imaging remains underexplored.

\subsection{Normalisation techniques}
Normalisation is a standard preprocessing step in computer vision that improves model training stability and performance by bringing image intensities into a consistent range \cite{lecunEfficientBackProp2012}. Two common normalisation techniques are Z-score and min-max scaling, which both apply linear transformations to image intensity values. In general computer vision tasks with natural images, Z-score normalisation typically uses dataset-wide mean and variance to rescale images. However, this approach is less suitable for MRI due to inter-image intensity variations. Instead, images are often transformed using image-specific means and standard deviations. Min-max scaling rescales images to a specified range, typically [0, 1]. This method is more sensitive to outliers than Z-score, so a more robust method of min-max scaling is often used, where image intensities are clipped to a given percentile before transformation, commonly [1st, 99th] percentiles \cite{islamEvaluationPreprocessingTechniques2021}.

However, these methods fail to account for acquisition variability across datasets or ensure that the same anatomical structures fall within consistent intensity ranges across images. Unlike imaging modalities such as CT, where pixel intensity has an intrinsic meaning which allows for targeted selection of tissue types (known as windowing), MRI exhibit large intensity variance across scans, making standardisation and quantitative analysis a challenge \cite{masoudiQuickGuideRadiology2021}. To address this, Ny\'{u}l and Udupa proposed a histogram matching method to better standardise MR image intensities. Introduced in \cite{nyulStandardizingMRImage1999} and refined in \cite{nyulNewVariantsMethod2000}, a histogram ``template'' is calculated from training image, and new images are transformed by matching histogram landmarks to this template. This method has become standard across MRI studies, and has been shown to be effective in various applications, such as identifying multiple sclerosis \cite{shahEvaluatingIntensityNormalization2011}. However, there are concerns that this method is not independent of biological variation and may therefore remove important information \cite{shinoharaStatisticalNormalizationTechniques2014}. That study, which focused on brain MRI, proposed white stripe normalisation, which uses a specific intensity peak (corresponding to normal-appearing white matter) for standardisation. However, no equivalent method exists for knee MRI, and other newly proposed MRI standardisation and harmonisation methods also focus solely on brain MRI \cite{wrobelIntensityWarpingMultisite2020,sedereviciusRobustIntensityDistribution2022,dalvitcarvalhodasilvaEnhancedPreprocessingDeep2022}.

MR images often suffer from low or non-uniform contrast, making it difficult to distinguish and segment specific structures. While normalisation methods such as Z-score and min-max scaling focus on standardising intensity distributions, they have little effect on the contrast of the image. Therefore, techniques for contrast enhancement are often used. Common approaches involve modification of the intensity histogram of an image \cite{hummelHistogramModificationTechniques1975}. Histogram equalisation (HE) improves the global contrast of an image by scaling the intensity distribution's cumulative distribution function, resulting in a more uniform histogram \cite{hummelImageEnhancementHistogram1977}. However, HE does not account for spatial intensity variations, which can lead to inadequate local contrast enhancement, as well as over- and under-enhancement in regions with intensity inhomogeneities. Adaptive histogram equalisation (AHE) addresses this limitation by applying HE locally \cite{pizerAdaptiveHistogramEqualization1987,hartingerAdaptiveHistogramEqualization2024}. However, this can over-amplify noise in homogeneous regions. Contrast-limited adaptive histogram equalisation (CLAHE) mitigates this issue by using a clipping limit to restrict enhancement in such regions \cite{hummelImageEnhancementHistogram1977}, and has been been shown to improve performance of deep learning models \cite{hartingerAdaptiveHistogramEqualization2024,hayatiImpactCLAHEbasedImage2023}.

\subsection{Related works}

Studies investigating the effect of intensity normalisation methods on the robustness of radiomics feature extraction from MRI have been performed, particularly in brain and breast imaging \cite{carreStandardizationBrainMR2020,liImpactPreprocessingHarmonization2021,schwarzhansImageNormalizationTechniques2025}. These studies found that intensity normalisation improved feature robustness, although the optimal method varied across specific tasks and datasets. However, they did not investigate deep learning tasks such as segmentation, and work on musculoskeletal imaging remains limited.

Few studies have systematically evaluated how preprocessing affects deep learning segmentation performance in MRI, especially on external datasets. One study tested external data only after model fine-tuning \cite{kondratevaNegligibleEffectBrain2024}, while another evaluated performance solely on internal data \cite{deraadEffectPreprocessingConvolutional2021}. Neither study found conclusive results, with only resampling providing noticeable improvements in performance \cite{deraadEffectPreprocessingConvolutional2021}. In contrast, a CT study found that windowing improved segmentation performance \cite{islamEvaluationPreprocessingTechniques2021}, but this method is not applicable to MRI due to its lack of intrinsic tissue-specific intensity values. To our knowledge, only one study has investigated the effect of MRI standardisation on segmentation performance when applied to data from other centres \cite{ghazvanchahiEffectIntensityStandardization2024}, but this focused on brain FLAIR MRI, with the best-performing method tailored to that modality. These studies highlight the limited understanding of how preprocessing impacts MRI segmentation performance, particularly under domain shift and outside of brain imaging.

A related study used the same two datasets as the present work to compare the performance of pretrained models on unseen datasets \cite{schmidtGeneralizabilityDeepLearning2023}. It found that sequence changes caused larger performance drops than scanner or centre effects. Findings emphasise the importance of addressing domain shift in medical imaging, especially when it involves different imaging protocols, and highlight the need to explore preprocessing strategies such as intensity normalisation that could help mitigate this issue.

\section{Materials and methods}

\subsection{Datasets}

\subsubsection{IWOAI 2019}

Data from the 2019 International Workshop on Osteoarthritis Imaging Knee Segmentation (IWOAI 2019) Challenge was used in this study for model training and internal testing \cite{desaiInternationalWorkshopOsteoarthritis2021}. This dataset is a subset of the Osteoarthritis Initiative (OAI), a multi-centre longitudinal study of patients at risk of femoral-tibial knee osteoarthritis \cite{OAIProtocol,peterfyOsteoarthritisInitiativeReport2008}. The IWOAI 2019 dataset contained 176 images composed of 3D sagittal double-echo steady-state (DESS) MRI of 88 patients at two time points (baseline and 1-year follow-up). Each image was size $384\times384\times160$ slices, with a spatial resolution of $0.36\text{mm}\times0.36\text{mm}\times0.7\text{mm}$ (slice thickness). Training, validation and test folds (of size 120, 24 and 24 respectively) were provided by challenge organisers, where Kellgren-Lawrence grade, BMI, and sex were approximately equally distributed across all folds. Manual segmentations of cartilage and menisci were also provided, annotated by a single expert segmenter from Stryker Imorphics \cite{desaiInternationalWorkshopOsteoarthritis2021}.

\subsubsection{SKM-TEA}

The Stanford Knee MRI with Multi-Task Evaluation (SKM-TEA) dataset was used for external testing \cite{desai2021skmtea}. It includes 3D quantitative double-echo steady-state (qDESS) MRI knee scans of 155 patients with knee pathology, acquired using two scanners at Stanford Healthcare. Images were size $512\times512\times2s$, with $s$ being the number of slices, which varied from 80 to 88 depending on knee size, and spatial resolution of $0.31\text{mm}\times0.31\text{mm}\times0.8\text{mm}$. The dataset included manual annotations of the lateral and medial meniscus, created by two researchers that were supervised by experienced musculoskeletal radiologists \cite{desai2021skmtea}. The qDESS sequence produces two separate images (``echos'') of different contrast \cite{schmidtGeneralizabilityDeepLearning2023,welschRapidEstimationCartilage2009}. SKM-TEA and IWOAI 2019 both used DESS-type sequences but employed different imaging protocols.

\subsection{Preprocessing}

For training, the images in the IWOAI 2019 dataset were cropped in the in-plane dimensions from $384\times384$ down to $200\times256$, with the approximately central cropping region selected based on the spatial distribution of ground truth segmentations in the training data. This reduced computational requirements during model training while increasing the proportion of the image containing the menisci. For external testing on the SKM-TEA dataset, the two echos of the qDESS sequence were first combined into a single image using the root-sum-of-squares method. Images were then cropped using the same width-to-height proportions as the training data, resulting in a final in-plane size of $266\times341$.

\subsection{Normalisation schemes}

Seven different data preprocessing methods were compared:

\renewcommand{\labelitemi}{\textbullet}
\begin{itemize}
    \item \textbf{Z-score}: Each image was standardised by its individual mean and standard deviation.
    \item \textbf{Min-max scaling}: Image intensities were linearly scaled to [0, 1].
    \item \textbf{Robust min-max scaling}: Images were clipped to the 1st and 99th percentiles before performing min-max scaling.
    \item \textbf{HE (histogram equalisation)}: The intensity histogram was equal\-ised globally across the entire image.
    \item \textbf{CLAHE (contrast-limited adaptive histogram equalisation)}: Image histograms were equalised locally using kernels of size 1/10th of the image dimensions. A clip limit of 0.01 was applied to prevent over-enhancement. HE and CLAHE were implemented using the scikit-image library \cite{waltScikitimageImageProcessing2014}.
    \item \textbf{Ny\'{u}l standardisation}: A histogram matching method proposed by Ny\'{u}l and Udupa \cite{nyulNewVariantsMethod2000}. A template was created using intensity landmarks from the training data. This template was then used to transform both the training and test data. The method was implemented using the torchIO package \cite{perez-garciaTorchIOPythonLibrary2021}.
    \item \textbf{GMM (Gaussian Mixture Model) normalisation}: Based on white stripe normalisation in brain MRI \cite{shinoharaStatisticalNormalizationTechniques2014}, this method standardises images using the mean of a selected intensity peak. A GMM with four components was fitted to the intensity distribution of each image. The component with the second largest mean was used to standardise the image.
\end{itemize}

\subsection{Model training}

The nnU-Net framework \cite{isenseeNnUNetSelfconfiguringMethod2021} was used to train all models. Although transformer-based architectures have become increasingly popular for medical image segmentation, nnU-Net remains a competitive state-of-the-art baseline, particularly in scenarios with limited training data \cite{isenseeNnUNetRevisitedCall2024}. Additionally, foundation models have been shown to underperform on meniscus segmentation \cite{millsPuttingSegmentAnything2025}, further supporting the use of a conventional but well-validated architecture. IWOAI 2019 training and validation splits were combined, and 3D U-Net models were trained using 5-fold cross-validation with identical splits across all normalisation methods. Experiments were planned using default nnU-Net settings \cite{isenseeNnUNetSelfconfiguringMethod2021}, resulting in a batch size of 2 and a patch size of 112$\times$128$\times$160. All models were trained for 500 epochs. During inference, we assessed model performance from each iteration of the cross-validation to evaluate the stability of the normalisation procedures.

\subsection{Metrics and statistical tests}

\subsubsection{Volumetric measures}

The Dice Similarity Coefficient (DSC), which measures the overlap of two sets, was used to assess volumetric agreement between the ground truth and predicted masks \cite{diceMeasuresAmountEcologic1945}. Variations in meniscal volumes between manual and automatic segmentations were assessed using Lin's concordance correlation coefficient (CCC) \cite{linConcordanceCorrelationCoefficient1989} and mean absolute error (MAE).

\subsubsection{Distance measures}

The Hausdorff Distance (HD) is a spatial distance metric that measures the maximum difference between closest points of two sets \cite{huttenlocherComparingImagesUsing1993}. It indicates how well spatial structures of the predicted segmentation align with the ground truth. HD is sensitive to outliers and noise, so this study used the 95th percentile Hausdorff Distance (HD95) for increased robustness \cite{tahaMetricsEvaluating3D2015}.

\subsubsection{Statistical analysis}

Due to large variation of DSC and HD95 scores across test images relative to inter-method differences, separate linear mixed-effects models (LMMs) were fitted to both metrics using predictions from all five fold-trained models for each method to investigate statistical significance while accounting for repeated measures within folds and images. LMMs were fitted using the \texttt{lme4} package in R, using the equation:

\begin{equation}
    \text{Metric} \sim \text{Method} + (1 | \text{Fold}) + (1 | \text{TestImage})
\end{equation}

with \textit{Method} as a fixed effect term and random intercept terms for \textit{Fold} and \textit{TestImage}. Z-score normalisation was used as the reference category for the fixed effect. An initial Type III analysis of variance (ANOVA) with Satterthwaite’s method was used to test for an overall effect of \textit{Method}. If significant, pairwise comparisons between methods were performed using Tukey-adjusted post hoc tests on the model estimates to control for multiple comparisons. A significance threshold of $\alpha = 0.05$ was used throughout.

\begin{table}[h]
    \small
    \centering
    \caption{Comparison of metrics across methods on internal (IWOAI 2019) and external (SKM-TEA) test data. Metrics are reported as the mean ± one standard deviation of values obtained from each of the five cross-validation model predictions. (HE: histogram equalisation; CLAHE: contrast-limited adaptive histogram equalisation; GMM: Gaussian Mixture Model)}
    \begin{tabular}{>{\raggedright\arraybackslash}p{0.21\textwidth}>{\centering\arraybackslash}p{0.17\textwidth}>{\centering\arraybackslash}p{0.16\textwidth}>{\centering\arraybackslash}p{0.18\textwidth}>{\centering\arraybackslash}p{0.12\textwidth}}
        \toprule
        \makecell{\textbf{Method}} & 
        \makecell{\textbf{DSC} \\ (\%) $\uparrow$} & 
        \makecell{\textbf{HD95} \\ (mm) $\downarrow$} & \makecell{\textbf{CCC} $\uparrow$} & 
        \makecell{\textbf{MAE} \\ (mm\textsuperscript{3}) $\downarrow$} \\
        \midrule
        \multicolumn{5}{c}{\textbf{Internal Test Data}} \\
        \midrule
        Z-score        & 89.04±0.12& 1.51±0.03& \textbf{0.939±0.006}& 225±9\\
        Min-max       & \textbf{89.05±0.07}& 1.50±0.03& 0.935±0.005& 230±14\\
        Robust min-max& 88.90±0.14& 1.52±0.04& 0.930±0.007& 252±15\\
        HE            & 88.68±0.09& 1.54±0.05& 0.926±0.008& 257±14\\
        CLAHE         & 88.83±0.10& 1.53±0.03& 0.928±0.016& 252±23\\
        Ny\'{u}l \cite{nyulNewVariantsMethod2000} & 88.93±0.19& 1.52±0.05& 0.927±0.014& 254±24\\
        GMM           & \textbf{89.05±0.14}& \textbf{1.48±0.05}& \textbf{0.939±0.005}& \textbf{224±14}\\
        \midrule
        \multicolumn{5}{c}{\textbf{External Test Data}} \\
        \midrule
        Z-score        & 78.95±0.28& 3.19±0.21& 0.829±0.008& 490±5\\
        Min-max       & 78.45±0.23& 3.21±0.13& 0.827±0.008& 499±6\\
        R. min-max& 78.48±0.26& 3.09±0.11& 0.819±0.010& 501±8\\
        HE            & 78.31±0.24& 3.30±0.15& 0.831±0.011& 491±5\\
        CLAHE         & 78.86±0.26& \textbf{3.05±0.23}& 0.831±0.005& 481±5\\
        Ny\'{u}l \cite{nyulNewVariantsMethod2000} & \textbf{79.10±0.34}& 3.19±0.24& \textbf{0.850±0.006}& \textbf{476±7}\\
        GMM           & 78.44±0.14& 3.17±0.27& 0.820±0.016& 502±9\\
        \bottomrule
    \end{tabular}
    \label{tab:results}
\end{table}

\section{Results}

\begin{figure}[h]
    \centering
    \includegraphics[width=0.81\linewidth]{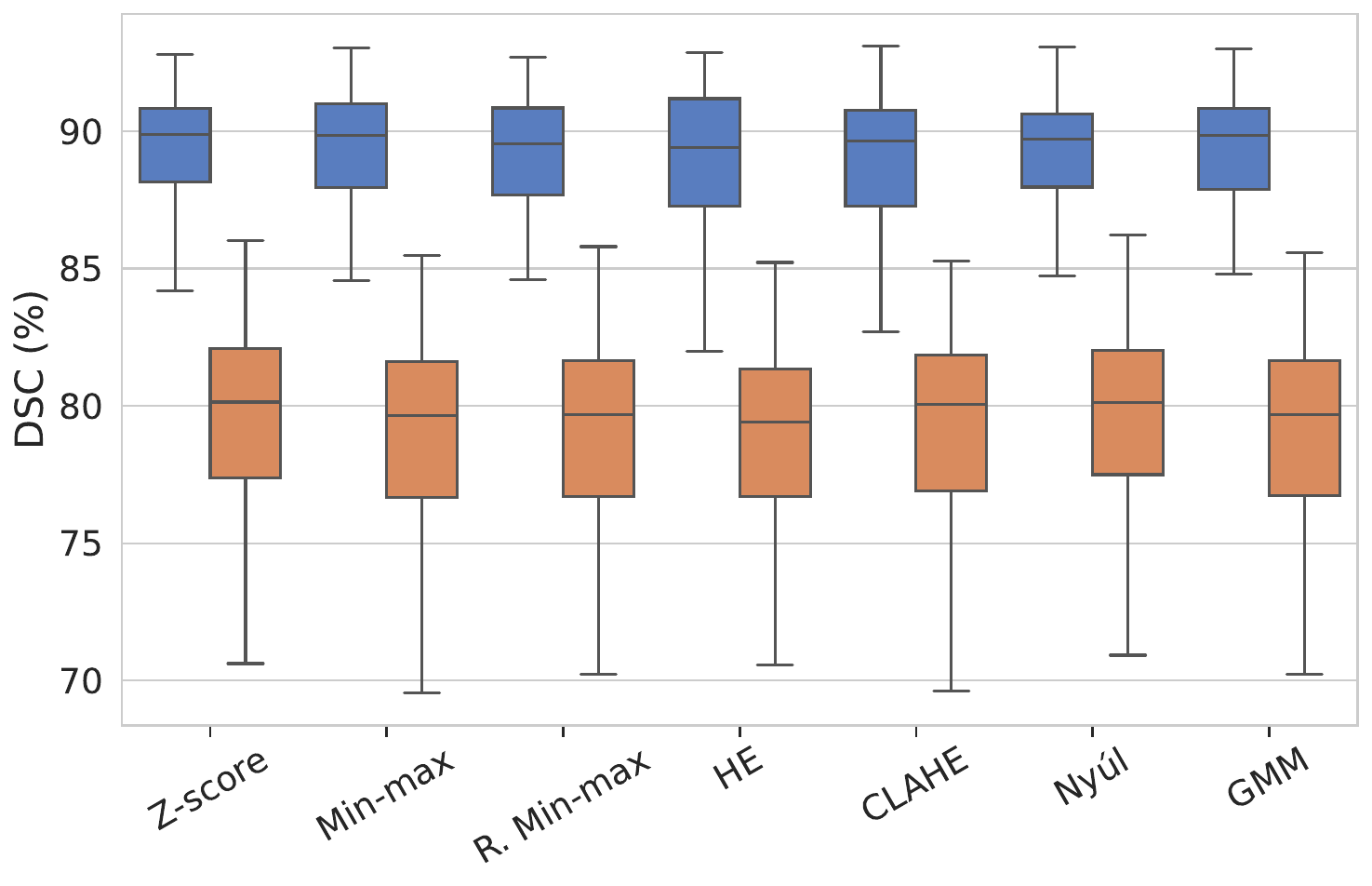}
    \caption{Box plots showing DSC performance of different preprocessing methods across test images on both internal (blue, IWOAI 2019) and external (orange, SKM-TEA) datasets. Outliers (identified using the 1.5×IQR rule) have been omitted for clarity, as their distribution were similar across methods.}
    \label{fig:all_dice}
\end{figure}

Models trained using different normalisation methods were evaluated on both internal (IWOAI 2019 test split) and external (SKM-TEA) data, with results shown in Table~\ref{tab:results}. Standard deviations reflect variation across the five fold-trained models. To visualise DSC distributions across test images, box plots are shown in Fig.~\ref{fig:all_dice}, where predictions from the five fold-trained models were ensembled by averaging voxel-wise softmax probabilities.

\subsection{Internal test set (IWOAI 2019)}
On the internal test set, min-max achieved the highest mean DSC, followed by GMM and Z-score. However, inter-method variation was small. A Type III ANOVA on the fitted LMM revealed a significant effect of method ($F(6,942)=8.08$, $p<0.001$), but pairwise comparisons found few statistically significant differences. The top three methods outperformed HE and CLAHE significantly, and HE performed significantly worse that all methods apart from CLAHE. Meanwhile, no significant effect of method was found for HD95 ($F(6,942)=1.71$, $p=0.12$). Comparison of manual vs. automated segmentation volumes (CCC and MAE) mirrored DSC results, with Z-score, min-max, and GMM performing best. HE was the lowest-performing preprocessing method across all metrics.

\begin{table}[!b]
\small
\centering
\caption{Estimated differences in DSC score compared to Z-score (reference category; estimated DSC = 78.95\%, CI [78.28, 79.63]) on external test data, based on a linear mixed-effects model with a fixed effect for method and random intercepts for fold and test image. Significance was assessed separately using Tukey-adjusted pairwise comparisons.}
\label{tab:mixed_effects_dsc}
\begin{tabular}{l>{\centering\arraybackslash}p{0.35\linewidth}>{\centering\arraybackslash}p{0.2\linewidth}}
\toprule
\textbf{Method} & \textbf{Estimated $\Delta$DSC (\%)} & \textbf{95\% CI} \\
\midrule
Min-max         & -0.51  & [-0.60, -0.41] \\
Robust min-max  & -0.48  & [-0.57, -0.38] \\
HE              & -0.65 & [-0.74, -0.55] \\
CLAHE           & -0.09  & [-0.19, 0.00] \\
Ny\'{u}l \cite{nyulNewVariantsMethod2000}  & \textbf{0.14}  & [0.05, 0.24] \\
GMM             & -0.51  & [-0.61, -0.42] \\
\bottomrule
\end{tabular}
\end{table}

\begin{figure}[ht]
    \centering
    \includegraphics[width=0.8\linewidth]{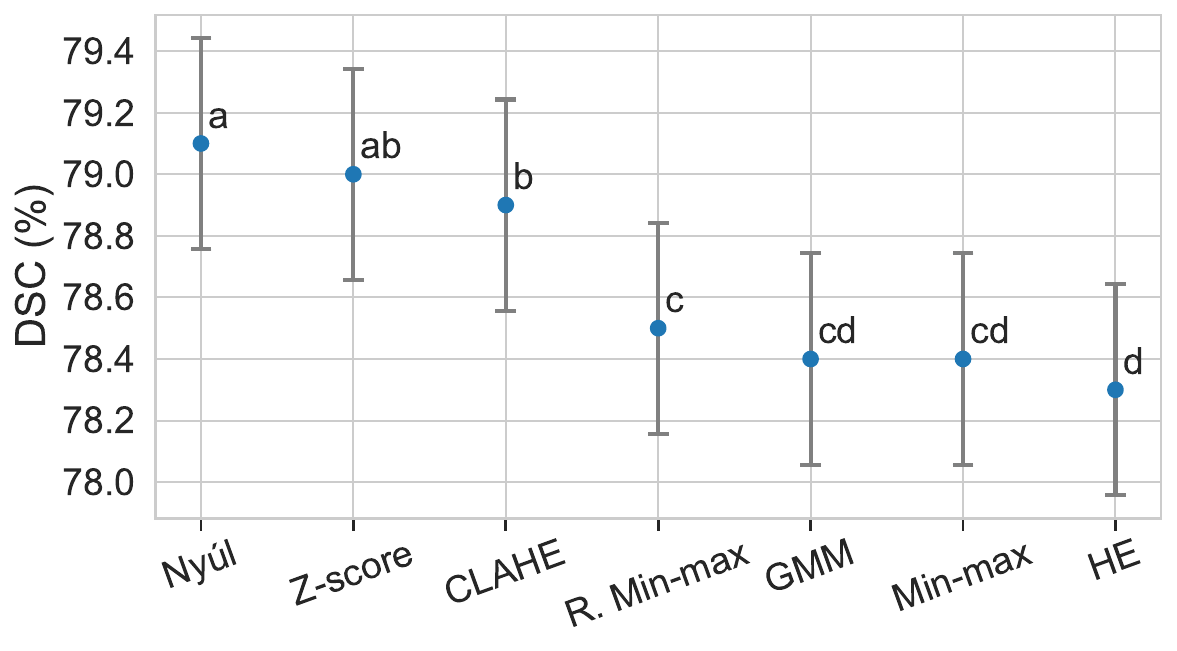}
    \caption{Estimated marginal means of DSC scores on the external test set (SKM-TEA) across normalisation methods, derived from a linear mixed-effects model. Error bars show 95\% confidence intervals. Letters next to each point represent compact letter display groups from Tukey-adjusted pairwise comparisons; methods sharing the same letter are not significantly different at the 5\% level.}
    \label{fig:lmm_dsc_emm}
\end{figure}

\begin{figure}[t]
    \centering
    \includegraphics[width=0.91\linewidth]{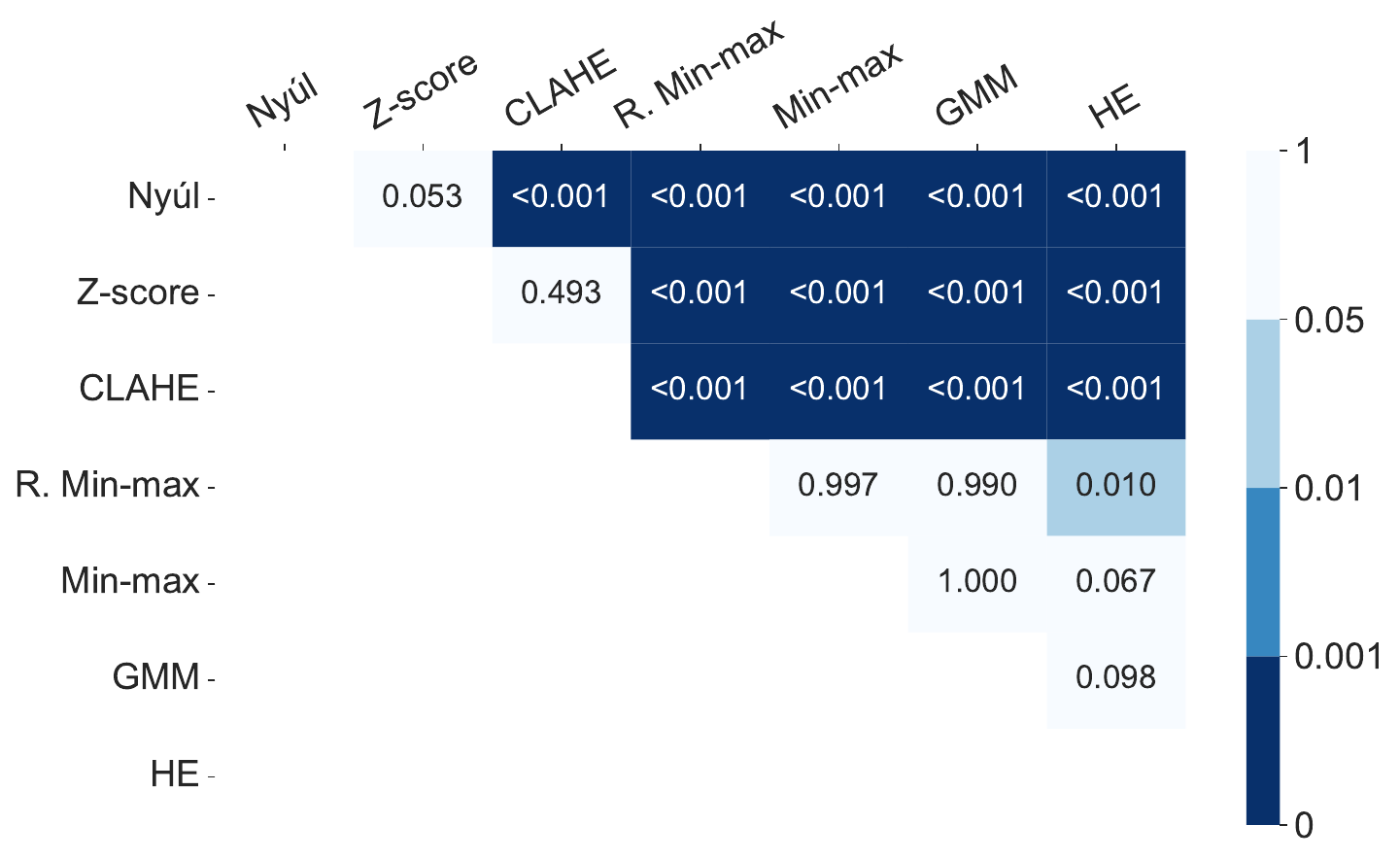}
    \caption{Tukey-adjusted pairwise \textit{p}-values from linear mixed-effects model comparisons of DSC scores between methods. Darker shades indicate greater statistical significance. Rows and columns are ordered by descending mean DSC.}
    \label{fig:p-values}
\end{figure}

\subsection{External test set (SKM-TEA)}

On the external test set, inter-method differences appeared more pronounced than in internal testing. Ny\'{u}l achieved the highest DSC, followed by Z-score and CLAHE. HE had the lowest DSC. Due to variation across test subjects concealing inter-method differences (Fig.~\ref{fig:all_dice}), an LMM was fitted to the DSC scores (Table~\ref{tab:mixed_effects_dsc}; Type III ANOVA: $F(6,5260)=79.06$, $p<0.001$). Variance decomposition indicated that most variability in DSC was attributable to differences between images (intra-class correlation coefficient (ICC) = 0.945), while the contribution from cross-validation folds was negligible (ICC = 0.003). The marginal $R^2$ was 0.005, indicating that normalisation method explained only 0.5\% of total variance. Estimated marginal means of DSC across the five methods, plotted in Fig~\ref{fig:lmm_dsc_emm}, show a separation between the top three and bottom four methods. Post hoc pairwise comparisons confirmed this, with Ny\'{u}l, Z-score, and CLAHE all outperforming min-max, robust min-max, HE, and GMM significantly ($p<0.001$). While Ny\'{u}l also performed significantly better than CLAHE ($p<0.001$), its advantage over Z-score was not statistically significant ($p=0.053$), despite the confidence interval for the fixed effect not containing zero (Table~\ref{tab:mixed_effects_dsc}). All pairwise \textit{p}-values are shown in Fig.~\ref{fig:p-values}. Ny\'{u}l also performed best on volumetric CCC and MAE. Although CLAHE achieved the lowest mean HD95, LMM analysis found no significant overall effect of method ($F(6,5264)=1.29$, $p=0.26$).

Among the two top-performing methods, Ny\'{u}l showed better volumetric consistency with ground truth annotations that Z-score, which tended to under-predict meniscus volume (mean difference: -103 mm$^3$ ($\sim1300$ voxels)). For a visual comparison of the two methods, image intensity profiles of both IWOAI 2019 and SKM-TEA datasets were plotted after transformation (Fig.~\ref{fig:intensities}). As seen in Fig.~\ref{fig:Z-score_hists}, differences between the datasets remain after Z-score, whereas Ny\'{u}l resulted in more consistent intensity profiles across the datasets (Fig.~\ref{fig:hist_standard}).

\section{Discussion}

Performance varied little on internal test data across normalisation methods, as expected given the shared domain with training images. Most methods performed comparably, with only HE and CLAHE performing significantly worse. This was likely due to their inconsistent contrast enhancement further disrupting tissue-specific intensity meaning across the dataset.

On the external test set, where domain shift was present, small but statistically significant differences between methods were observed, with LMM analysis identifying statistically significant advantages for Ny\'{u}l, Z-score, and CLAHE. Inter-method variation was small relative to variation between images (Fig.~\ref{fig:all_dice}), highlighting the difficulty of this segmentation task. Differences between methods were modest ($\sim1\%$ DSC), whereas all methods experienced a substantially larger performance drop between internal and external datasets ($\sim10\%$ DSC), indicating that domain shift has a much greater impact on performance than the choice of normalisation method. Notably, the marginal $R^2$ indicated that normalisation method explained only a small proportion of total variance. No significant differences were seen in HD95, indicating that segmentation shape was consistent across methods.

\begin{figure}[t]
\centering
\subfloat[Z-score Normalisation\label{fig:Z-score_hists}]{%
  \includegraphics[width=0.49\textwidth]{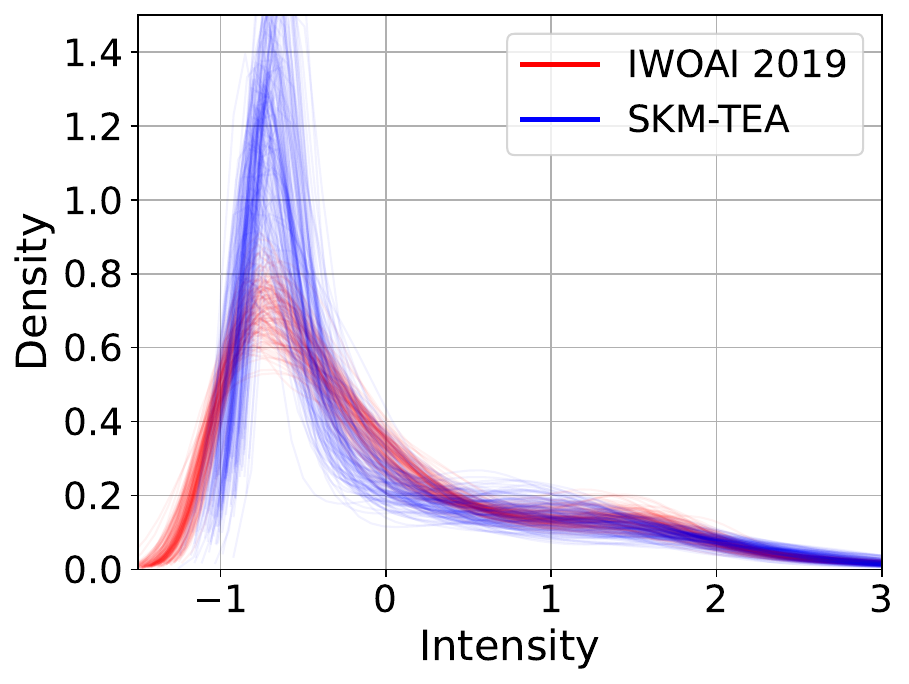}%
}\hfil
\subfloat[Ny\'{u}l Histogram Matching \cite{nyulNewVariantsMethod2000}\label{fig:hist_standard}]{%
  \includegraphics[width=0.49\textwidth]{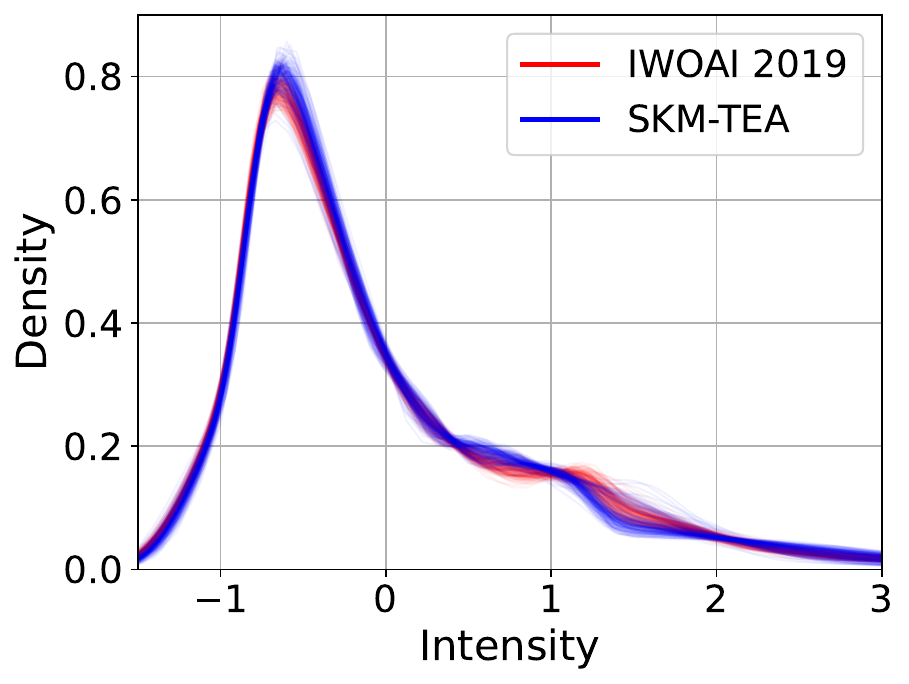}%
}
\caption{Comparison of intensities of images from both the IWOAI 2019 (red) and SKM-TEA (blue) datasets after different normalisation methods.}
\label{fig:intensities}
\end{figure}

\sloppy{
Differences between internal and external performance revealed method-specific behaviour. Min-max scaling suffered the largest performance drop, likely due to sensitivity to dataset-specific intensity outliers, whereas Ny\'{u}l showed the smallest drop, suggesting greater robustness to cross-dataset intensity  variations. CLAHE, which showed lower performance on the internal test set, ranked among the top methods externally, suggesting that contrast alteration might offer some generalisability benefits under domain shift. In contrast, the GMM-based approach performed well internally but less well externally, likely due to differences in intensity profiles between datasets affecting peak estimation (Fig.~\ref{fig:Z-score_hists}).}

\begin{figure}[!t]
\centering
\subfloat[Ground Truth\label{fig:groundtruth_eg}]{%
  \includegraphics[width=0.24\textwidth]{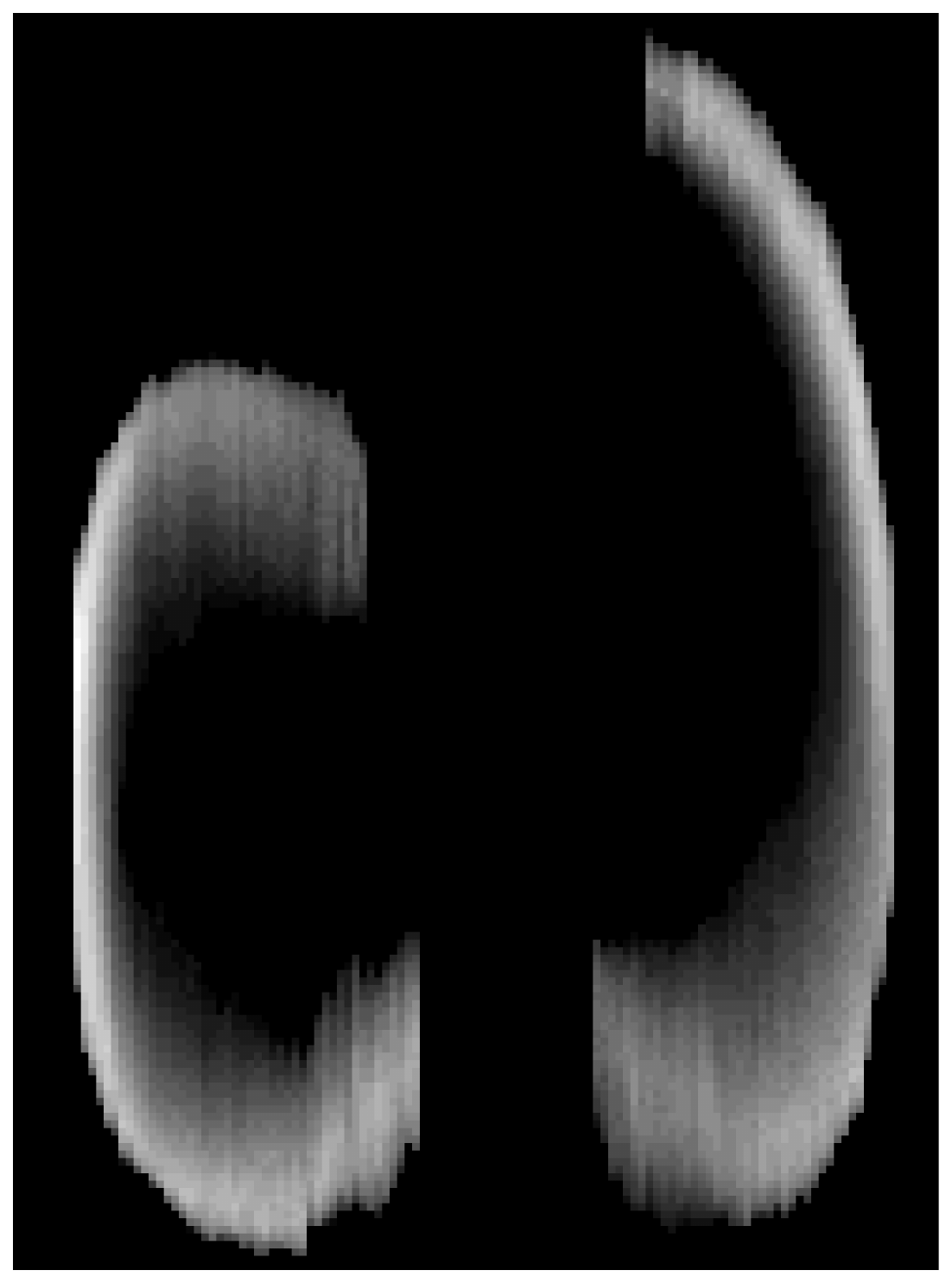}%
}\hfil
\subfloat[Z-score\label{fig:zscore_eg}]{%
  \includegraphics[width=0.24\textwidth]{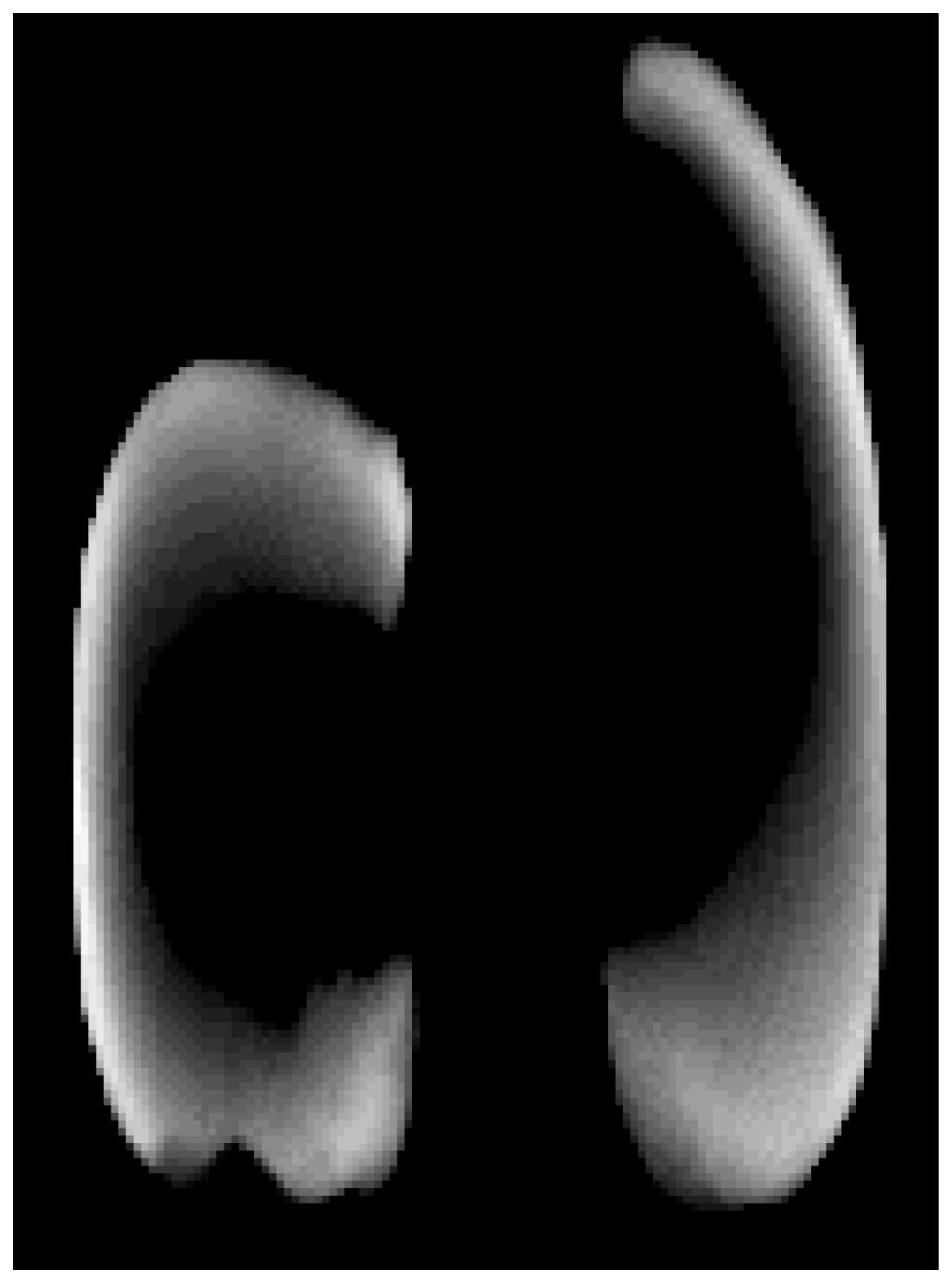}%
}\hfil
\subfloat[Min-max\label{fig:minmax_eg}]{%
  \includegraphics[width=0.24\textwidth]{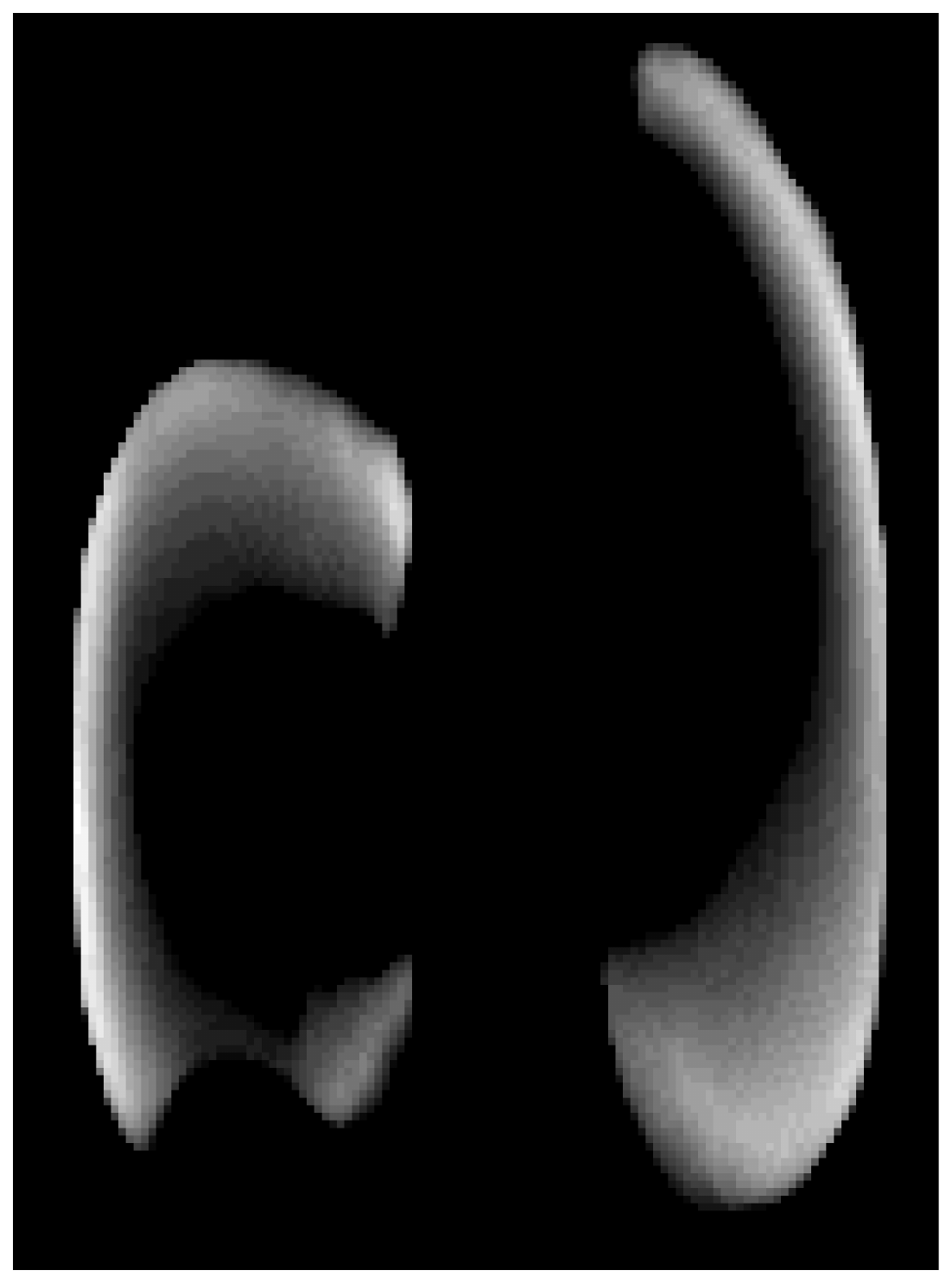}%
}\hfil
\subfloat[Robust min-max\label{fig:r_minmax_eg}]{%
  \includegraphics[width=0.24\textwidth]{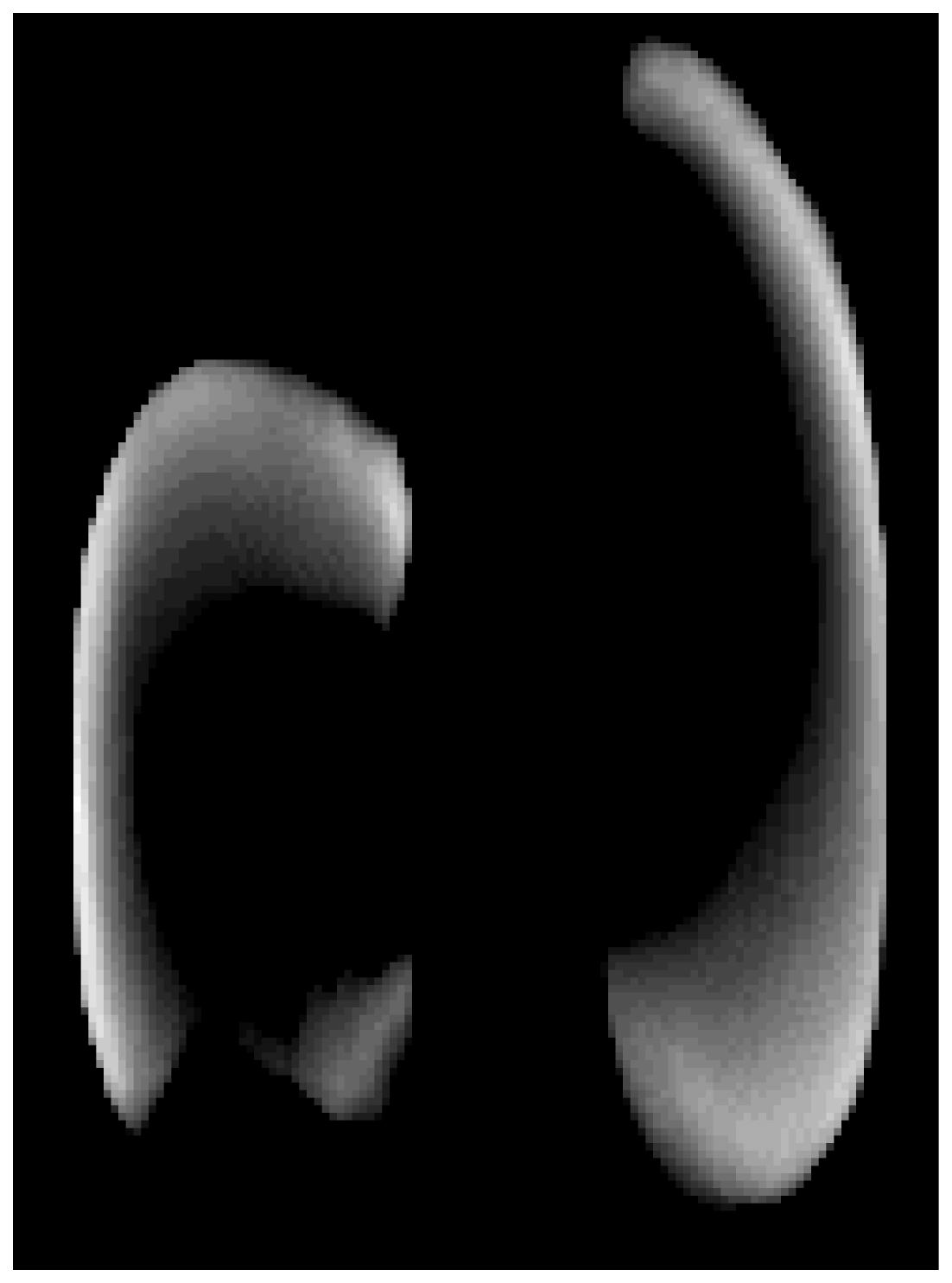}%
}\vfil
\subfloat[HE\label{fig:he_eg}]{%
  \includegraphics[width=0.24\textwidth]{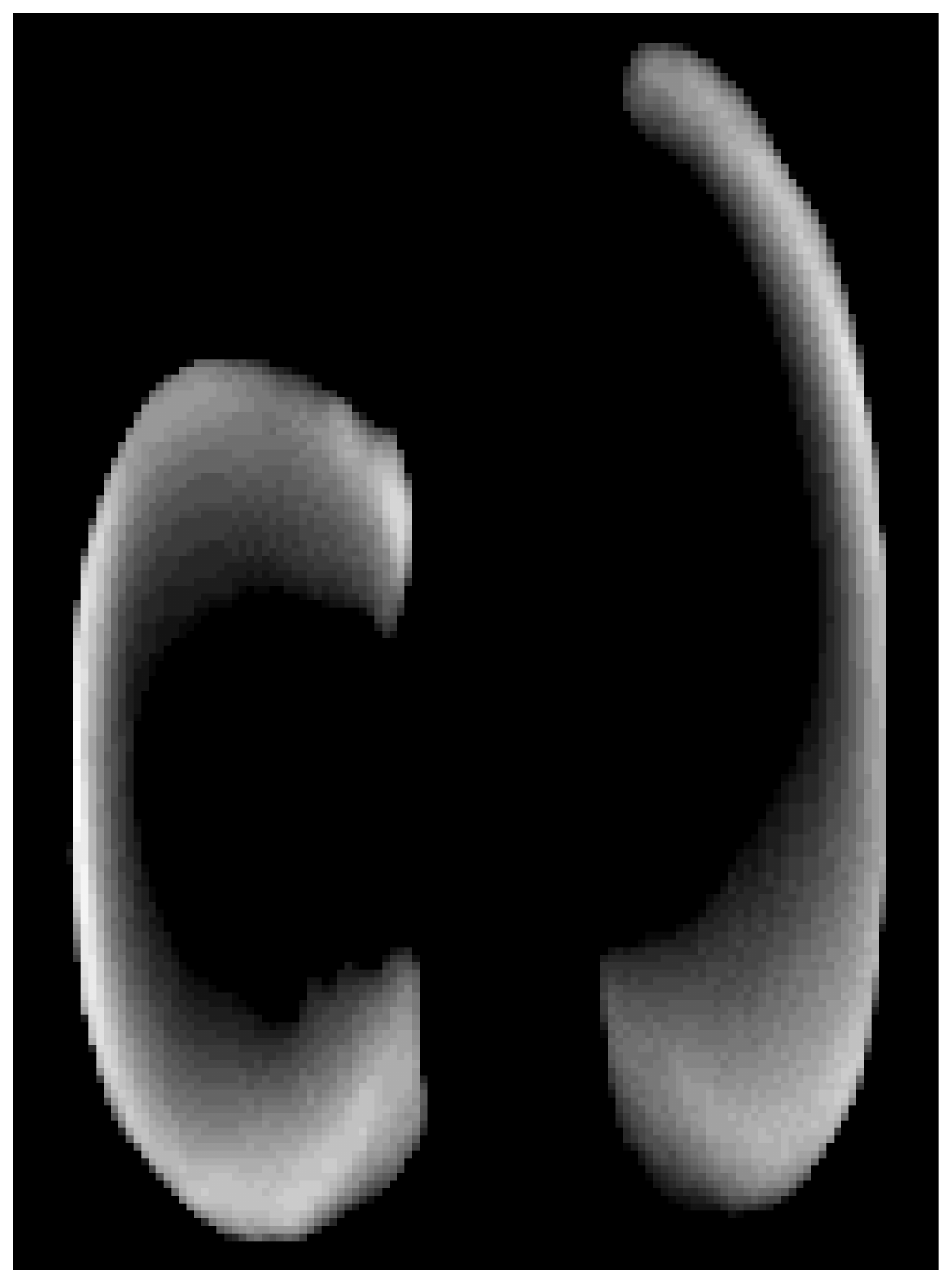}%
}\hfil
\subfloat[CLAHE\label{fig:clahe_eg}]{%
  \includegraphics[width=0.24\textwidth]{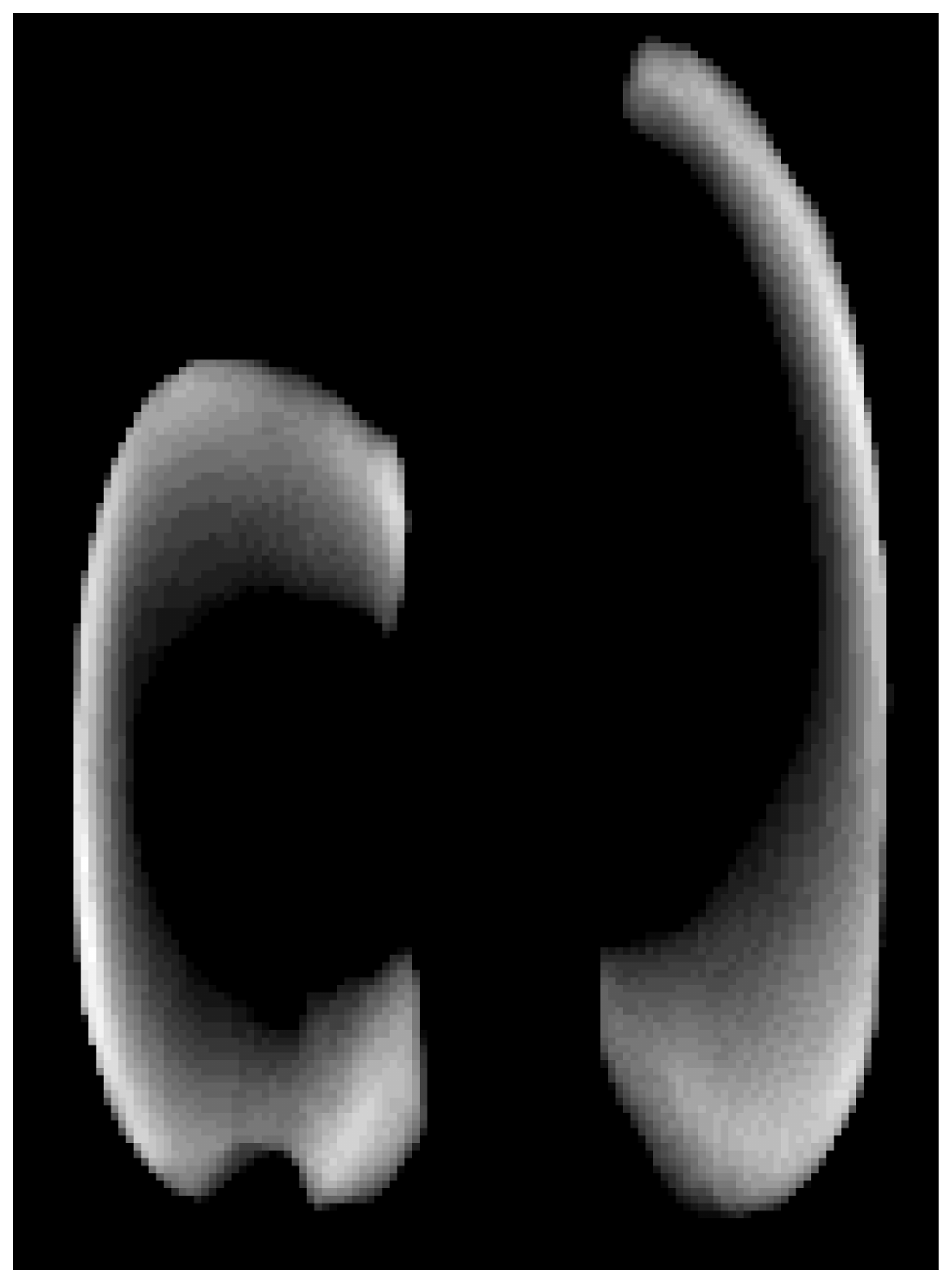}%
}\hfil
\subfloat[Ny\'{u}l \cite{nyulNewVariantsMethod2000}\label{fig:nyul_eg}]{%
  \includegraphics[width=0.24\textwidth]{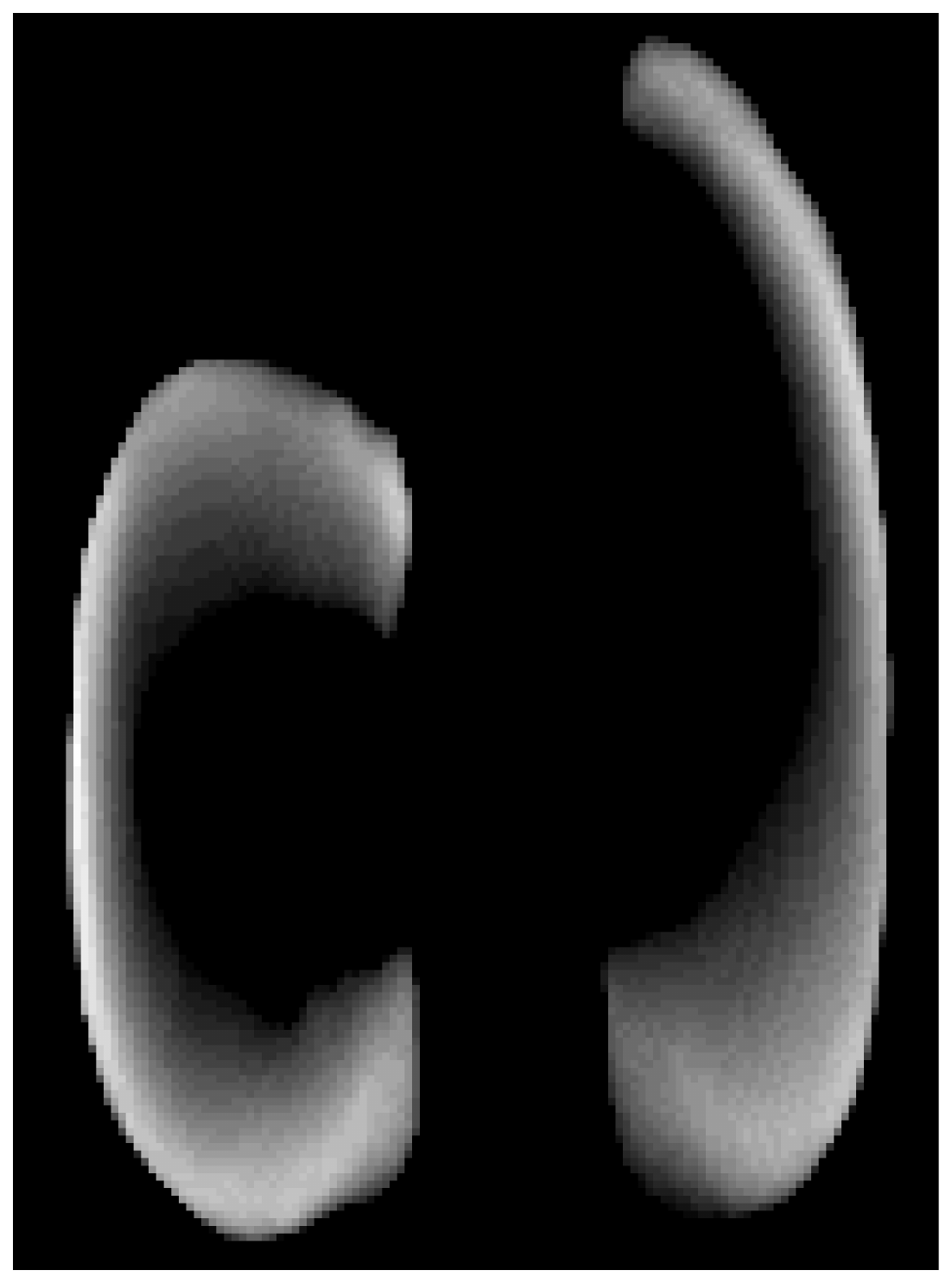}%
}\hfil
\subfloat[GMM\label{fig:gmm_eg}]{%
  \includegraphics[width=0.24\textwidth]{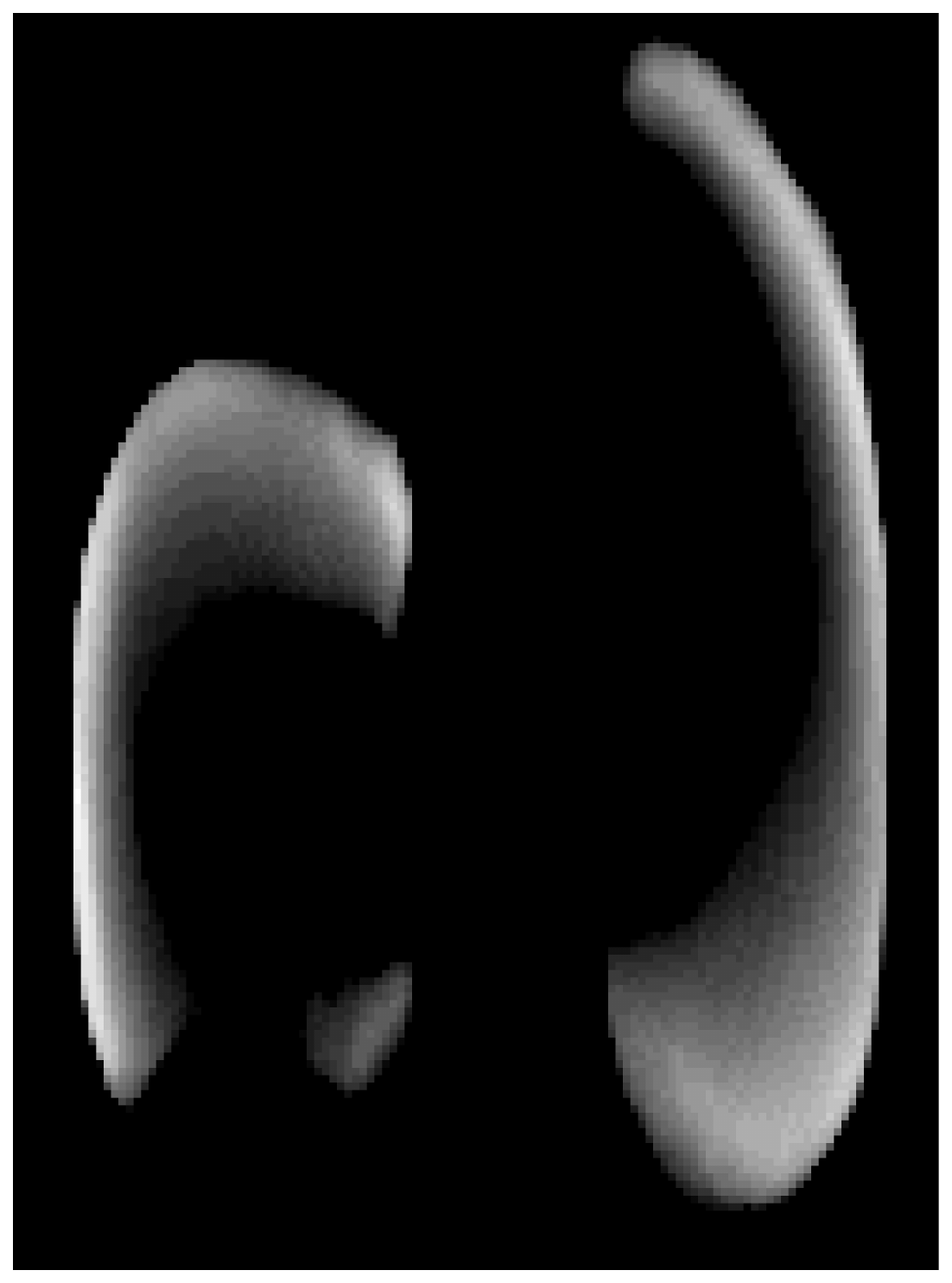}%
}
\caption{Qualitative comparison of meniscus segmentation across normalisation methods for the case with the greatest variability in Dice score. (a) Ground truth; (b–h) predictions from seven methods. Masks are shown as 2D representations, created by summing the 3D masks along the inferior-superior axis. A notable difference is seen in the lateral posterior horn (bottom left of each panel), where several methods (e.g., min–max, robust min–max, and GMM-based normalisation) introduce a gap not present in the ground truth.}
\label{fig:example}
\end{figure}

Among the top-performing methods, Z-score slightly underestimated meniscal volumes, while Ny\'{u}l had minimal volumetric bias. Although the magnitude of the underestimation was small ($\sim2\%$ of total volume), the majority of these inaccuracies fell on the meniscal boundaries, which could affect surface features. An example is shown in Fig.~\ref{fig:example}, where several methods introduce a  gap in the lateral posterior horn that is not present in the ground truth; notably, this artifact is not observed with Ny\'{u}l normalisation.

Despite Ny\'{u}l achieving better alignment of intensity distributions compared to Z-score (Fig.~\ref{fig:intensities}), this did not translate to better segmentation performance. This suggests that histogram matching does not necessarily improve performance, and could come at the cost of preserving biological variation.

Differences between datasets likely contributed to the observed performance drop. Although both datasets used DESS-type MRI sequences, their imaging protocols differed (e.g. echo time, repetition time), leading to subtle differences in image contrast (Fig.~\ref{fig:Z-score_hists}). Patient populations also differed: IWOAI 2019 includes patients at risk of or with early-stage osteoarthritis (OA), whereas SKM-TEA specifically recruited patients with knee pathologies, likely including more severe, late-stage OA cases. However, no significant difference in DSC was observed between images with and without recorded pathology labels in the external dataset (Mann–Whitney U test, $p=0.67$), suggesting that pathology presence alone does not explain the observed performance differences. Annotation differences are another source of variation. On a challenging task such as meniscus segmentation, inter-annotator variability could be large, but this is difficult to assess and is a limitation of this study. However, the lack of systematic over- or under-segmentation of volume suggests no major bias. 

Advanced harmonisation and domain adaptation methods, including adversarial approaches, style transfer, and ComBat-based techniques, have been proposed to address domain shift in medical imaging. These approaches often rely on access to target-domain data or task-specific adaptation. In contrast, this study evaluates simple intensity normalisation methods that require no target-domain information, providing a practical baseline for comparison.

\section{Conclusion}

This study evaluated the impact of different intensity normalisation methods on knee MRI meniscus segmentation, both internally and externally. Internal performance differences were minimal, with all methods performing similarly when trained and tested within the same domain. On external data, fold-level analysis with linear mixed effect modelling revealed small but statistically significant differences, with Ny\'{u}l, Z-score, and CLAHE outperforming other methods on volume-based metrics. However, these differences were modest compared to the performance drop observed between internal and external data. This suggests that, while normalisation choice can have a measurable effect on model generalisability, its contribution is limited relative to the impact of domain shift.

\subsection*{Data availability}
The datasets used in this study are publicly available. IWOAI 2019 can be accessed upon request at  
\url{https://github.com/denizlab/2019_IWOAI_Challenge} while SKM-TEA is available at 
\url{https://doi.org/10.71718/2ghb-nv62}, with information and tutorials at \url{https://github.com/StanfordMIMI/skm-tea}. All code developed for preprocessing, training, and evaluation is publicly available at \url{https://github.com/oliverjm1/mri_normalisation}.

\subsection*{Acknowledgements}
We would like to thank the OAI and its participants for creating this publicly available data set, Dr Akshay Chaudhari (Stanford University) for providing the subset used in the IWOAI 2019 challenge, and the participants of the SKM-TEA dataset for their valuable contributions. This work was undertaken on Aire, part of the High Performance Computing facilities at the University of Leeds, and was funded by EPSRC (EP/S024336/1).

\bibliographystyle{splncs04}
\bibliography{bib/preprocPaperRefs,bib/extrarefs}

\end{document}